\documentclass{article}
\usepackage{spconf,amsmath,graphicx}
\usepackage{booktabs}    
\usepackage{multirow}    
\usepackage{makecell}    
\usepackage{tabularx} 

\title{Oculomix: Hierarchical Sampling for Retinal-Based Systemic Disease Prediction}
\name{
\begin{tabular}{@{}c@{}}
Hyunmin Kim$^{1,2}$ \qquad 
Yukun Zhou$^{1,2}$ \qquad 
Rahul A. Jonas$^{1,2,3}$\qquad 
Lie Ju$^{1,2}$ \\
Sunjin Hwang$^{4}$ \qquad 
Pearse A. Keane$^{1,2}$ \qquad 
Siegfried K. Wagner$^{1,2}$ 
\end{tabular}
}
\address{
    $^{1}$Institute of Ophthalmology, University College London, United Kingdom\\
$^{2}$NIHR Moorfields Biomedical Research Centre, London, United Kingdom\\
$^{3}$University of Cologne, Faculty of Medicine, Department of Ophthalmology, Germany\\
$^{4}$Hanyang University Guri Hospital, Guri City, South Korea
}
\begin{document}
%
\maketitle
\begin{abstract}
Oculomics — the concept of predicting systemic diseases, such as cardiovascular disease and dementia, through retinal imaging — has advanced rapidly due to the data efficiency of transformer-based foundation models like RETFound.
Image-level mixed sample data  augmentations, such as CutMix and MixUp, are frequently used for training transformers, yet these techniques perturb patient-specific attributes, such as medical comorbidity and clinical factors, since they only account for images and labels.

To address this limitation, we propose a hierarchical sampling strategy, \textbf{Oculomix}, for mixed sample augmentations. Our method is based on two clinical priors. First (exam level), images acquired from the same patient at the same time point share the same attributes. Second (patient level), images acquired from the same patient at different time points have a soft temporal trend, as morbidity generally increases over time. Guided by these priors, our method constrains the mixing space to the patient and exam levels to better preserve patient-specific characteristics and leverages their hierarchical relationships.
The proposed method is validated using ViT models on a five-year prediction of major adverse cardiovascular events (MACE) in a large ethnically diverse population (Alzeye). We show that Oculomix consistently outperforms image-level CutMix and MixUp by up to 3\% in AUROC, demonstrating the necessity and value of the proposed method in oculomics.
\end{abstract}

\begin{figure}[t]  
    \centering
    \includegraphics[width=1.0\linewidth]{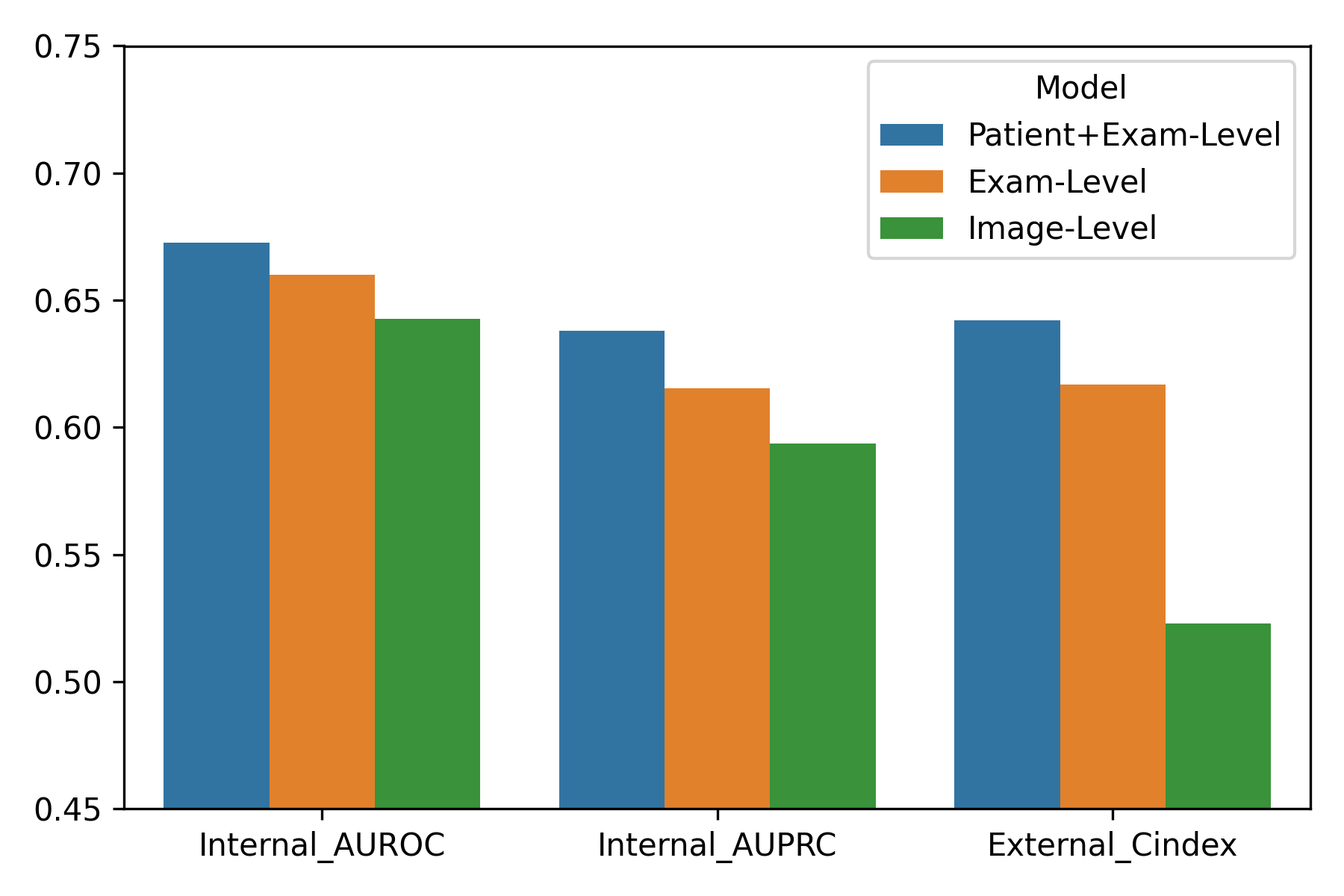}
    \caption{Internal (AlzEye test set; AUROC, AUPRC) and external (HYU MACE dataset; C-index) performance of average of ViT-S and ViT-B with CutMix+MixUp under three sampling strategies: Image-level (baseline), Exam-level, and Patient+Exam-level.}
    \label{fig:loss_auc}
\end{figure}

\begin{figure*}[t] 
    \centering
    \includegraphics[width=\textwidth]{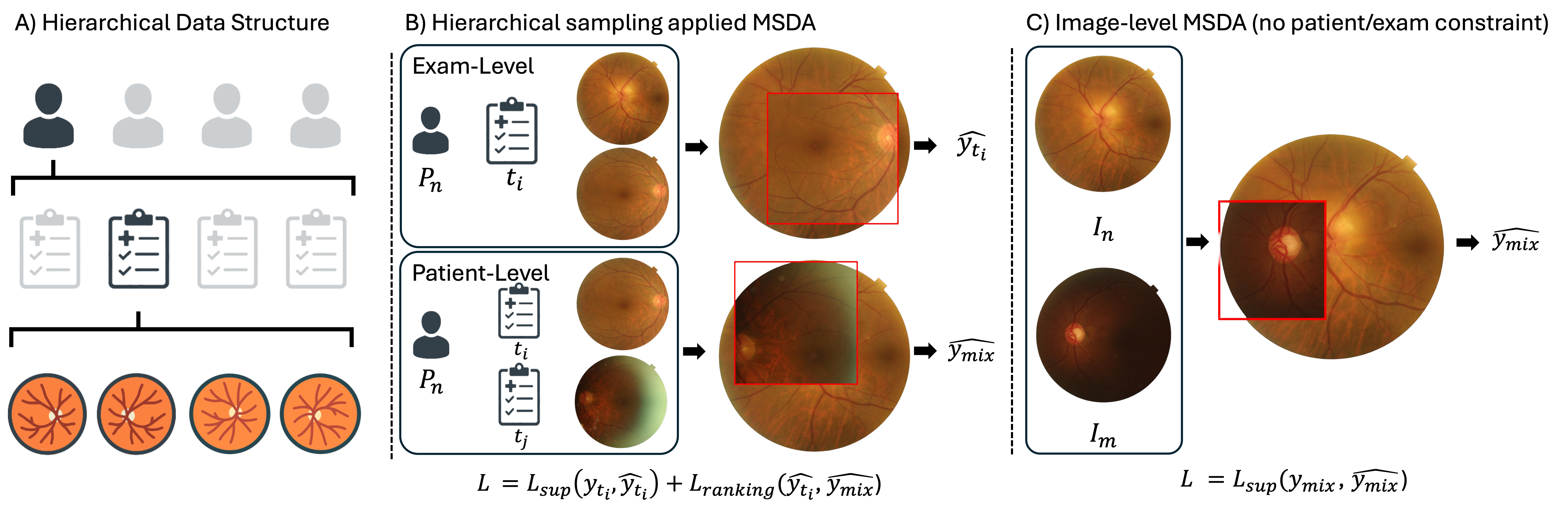}
    \caption{Overview of the proposed sampling strategy for mixed sample data augmentation (MSDA). 
\textbf{(A)} Hierarchical longitudinal ophthalmic datasets include patient-, exam-, and image-level components. 
\textbf{(B)} In hierarchical MSDA, a pairwise ranking loss captures the temporal progression between exams. 
$P_n$ denotes a patient, and $t_i$ and $t_j$ represent different exam time points. 
The prediction from mixed images at the same time point is denoted as $\hat{y}_{t_i}$, with $y_{t_i}$ as the corresponding label, 
while $\hat{y}_{\text{mix}}$ represents the prediction from mixed images across time points $t_i$ and $t_j$. 
$L_{\text{sup}}$ and $L_{\text{ranking}}$ represent supervised and ranking losses, respectively. 
\textbf{(C)} In image-level MSDA, arbitrary images are mixed regardless of patient identity or exam time point, 
with $y_{\text{mix}}$ and $\hat{y}_{\text{mix}}$ denoting the mixed label and its prediction.
}
    \label{fig:overall}
\end{figure*}

\section{Introduction}
\label{sec:intro}
Oculomics \cite{siggy_oculomics} is the field of predicting systemic health and disease biomarkers using retinal images in ophthalmology. Substantial research \cite{zhu_oculomics} enabled by machine learning and accumulated clinical data has demonstrated its potential across various systemic diseases \cite{siggy_parkinson} including cardiovascular disease, dementia, and chronic kidney disease.

However, a scarcity of well-labeled clinical data remains a major challenge. Specialist equipment, such as computed tomography (CT) scanners are costly, expert annotation is time consuming, and reliable longitudinal follow-up is often lacking. Therefore, foundation models  \cite{retfound} trained on a large amount of unlabeled data have been highlighted as a promising solution to mitigate data scarcity in Oculomics research. For building foundation models, vision transformers \cite{vit} are widely adopted due to their versatile and scalable architecture, but their lack of inductive biases in vision transformers complicates their training. Consequently, identifying the optimal training schemes\cite{deit} for vision transformers is an active area of research.

Numerous studies have shown that mixed sample data augmentations (MSDA), such as CutMix\cite{cutmix} and MixUp\cite{mixup}, are crucial for effective training with improvements not only in generalizability and robustness but also in pixel level regularization. As a result, these approaches have become a de facto standard in training vision transformers.

Yet, existing mixed sample data augmentation techniques, including extensions of previous work such as PuzzleMix\cite{puzzlemix} and TransMix\cite{transmix}, only considers class labels such as the presence or absence of disease neglecting clinical factors, such as \cite{qrisk} family history, genetics, lifestyle and socioeconomic status - key determinants for understanding, diagnosing and characterizing systemic disease. Therefore, such mixing approaches can perturb these critical characteristics, leading to the loss of clinically meaningful information that reflects individual morbidity.

To prevent perturbing patient-specific attributes, we introduce a hierarchical sampling strategy reflecting patient-exam-image hierarchy, in contrast to the canonical mixed sample data augmentation that operates only at the image level. This strategy is grounded in two clinical priors based on the inherent structure of the clinical dataset. First, at the exam level, multiple images captured at the same time point, typically representing a single visit and including both eyes with macular and disc-centered views, share a common clinical context. Second, at the patient level, longitudinal exams from the same patient reflect gradual changes, according to the progressive accumulation of morbidity over time.

Building on these priors, the scope of data mixing at the exam and patient levels is constrained. At the exam level, the scope of data mixing is restricted within the same exam to preserve shared attributes, including class labels and clinical factors. At the patient level, the scope of mixing is similarly limited to the same patient, enabling diverse combinations across different exams while maintaining patient identity. Although mixed labels across different exams cannot be strictly defined, we address this by leveraging their temporal order as an alternative form of supervision.

We evaluate our method on the task of five-year MACE prediction in the AlzEye \cite{alzeye} dataset. Our experiments show consistent improvements across all hierarchical levels, by achieving a performance improvement of 3.0\% in AUROC and 4.4\% AUPRC. Furthermore, our proposed method demonstrates superior performance in survival analysis on an independent external test set (the HYU MACE dataset) with a 12\% improvement in C-index. We summarize our contributions as two major strands:

\begin{itemize}
    \item We propose a hierarchical sampling strategy that leverages the patient–exam–image hierarchy, thereby preserving and fully utilizing patient-specific attributes while reflecting real clinical scenarios.
    \item Based on the standard augmentations CutMix and MixUp, we comprehensively validate our sampling method, demonstrating consistent improvements across internal and external datasets.
\end{itemize}

\section{method}
\label{sec:format}

\subsection{Exam-level sampling}
\label{ssec:subhead}

In contrast to conventional image classification tasks, such as cats vs. dogs, where classes are clearly defined, systemic diseases are complex, multifactorial, and influenced by the interaction of multiple clinical factors\cite{qrisk}, including age, sex, genetics, family history, and social deprivation. From this perspective, blending only systemic disease labels during mixed sample data augmentation can oversimplify and perturb these inherently complex relationships.

To address this issue, we propose exam-level sampling. Unlike conventional mixed sample data augmentation, which freely combines images and blends labels, exam-level sampling restricts mixing to images captured at the same time point from a single patient —typically including both eyes with macula- and disc-centered views— since these images share consistent clinical attributes such as age, sex, and systemic disease labels.

\begin{table}[t]
    \centering
    \setlength{\tabcolsep}{4pt} 
\begin{tabular}{llcc|c}
    \toprule
    Model & Sampling & \multicolumn{2}{c|}{Internal} & External \\
    \cmidrule(lr){3-4} \cmidrule(lr){5-5}
          &        & AUROC & AUPRC & C-index \\
    \midrule
    \multirow{3}{*}{ViT-Small} 
        & Patient+Exam & \textbf{67.22} & \textbf{63.79} & \textbf{63.43} \\
        & Exam         & 65.38 & 60.46 & 61.07 \\
        & Image     & 63.54 & 58.25 & 54.95 \\
    \midrule
    \multirow{3}{*}{ViT-Base} 
        & Patient+Exam & \textbf{67.29} & \textbf{63.81} & \textbf{64.98} \\
        & Exam         & 66.62 & 62.60 & 62.32 \\
        & Image     & 64.99 & 60.49 & 49.62 \\
    \bottomrule
\end{tabular}
    \caption{Comparison of ViT model performance (\%) across sampling strategies with CutMix+MixUp augmentation on internal and external test sets.}
    \label{tab:results}
\end{table}

\subsection{Patient-level sampling}
\label{ssec:subhead}

To expand the range of candidate mixing combinations without perturbing patient identity, we introduce patient-level sampling, which incorporates images from different exams of the same patient. However, defining a mixed label when combining images from different exams of the same patient is inherently ambiguous, because systemic disease progression is continuous, whereas diagnostic labels are typically binarized, making it difficult to reflect this property in the mixing process.

Therefore, we leverage their temporal order as an alternative label to reflect the relationship between morbidity and time, rather than directly using a mixed label. We implement this as a soft constraint using pairwise ranking loss~\cite{pairwise}, enabling mixtures of images from different exams to receive relative supervision by comparison with mixtures from the same exam. This loss is formulated as follows:

\begin{equation*}
\mathcal{L}_{\text{ranking}}(l_{t_1},l_{t_2}) =
\begin{cases}
\operatorname{ReLU}\!\big(m - (l_{t_2}-l_{t_1})\big), & t_1\!<\!t_2,\\[-2pt]
\operatorname{ReLU}\!\big(m - (l_{t_1}-l_{t_2})\big), & t_1\!>\!t_2.
\end{cases}
\end{equation*}
\noindent
where $m$ denotes the margin, $l$ represents the predicted logit, and $t$ indicates the exam time point.

In summary, our method consists of two steps: exam-level sampling, which combines images from the same patient at the same time point, and patient-level sampling, which combines images from the same patient at different time points.  In exam-level sampling, supervision is provided by the label associated with a single exam, whereas relative supervision is received in patient-level sampling. As a result, our approach is trained with the following objective function, as described in Figure 2:
\[
\mathcal{L} = \mathcal{L}_{\text{sup}}\!\left(y_{t_i}, \hat{y}_{t_i}\right)
+ \mathcal{L}_{\text{ranking}}\!\left(\hat{y}_{t_i}, \hat{y}_{\text{mix}}\right)
\]

\begin{figure}[t]
    \centering
    \includegraphics[width=\linewidth]{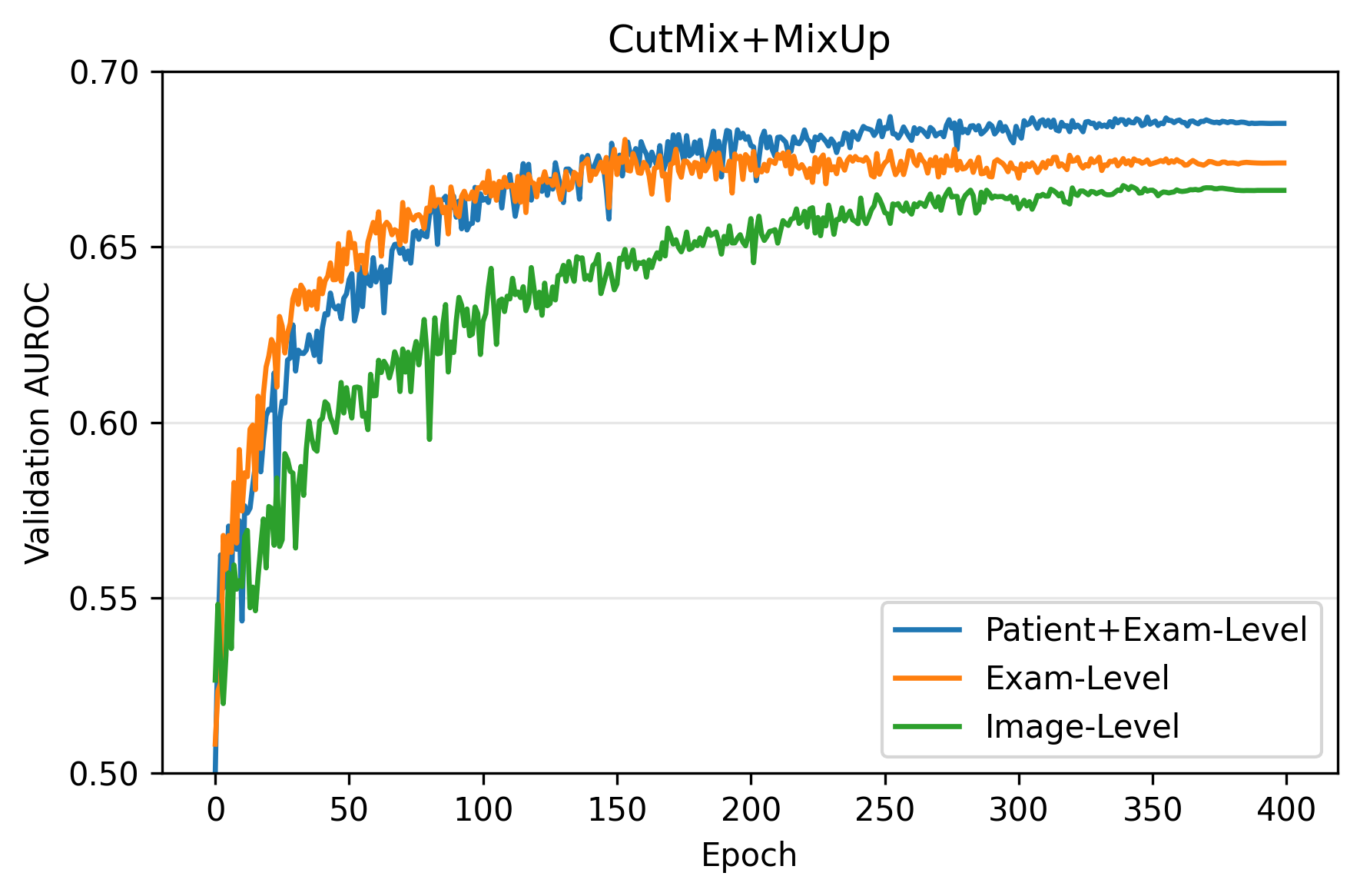}
    \caption{AUROC curves on the AlzEye validation dataset using CutMix+MixUp across sampling strategies.}
    \label{fig:training_graph}
\end{figure}

\section{experiments}
\label{sec:pagestyle}

\subsection{Datasets}
\label{ssec:setup}

Our proposed method is trained and evaluated on the AlzEye dataset, which contains longitudinal retinal imaging linked to 5-year major adverse cardiac events (MACE). The dataset is partitioned into training, validation, and test sets comprising 9,366, 3,746, and 2,812 patients; 15,246, 6,026, and 5,209 exams; and 29,475, 11,513, and 10,984 images, respectively. In the training dataset, 36.50\% of patients undergo two or more exams, and 44.46\% of exams include two or more images. For external test, we utilize the HYU MACE dataset, which is a privately curated cohort of 1,066 patients from HYU Hospital in South Korea under the same criteria as AlzEye.

\subsection{Experiment setup}
\label{ssec:setup}

We train ViT-Base and ViT-Small using the AdamW optimizer with a learning rate of 2e-4, a batch size of 512, label smoothing of 0.1, cosine learning rate decay, and a stochastic depth rate of 0.2, for 400 epochs at a resolution of 384 × 384. The cross-entropy loss is utilized as the supervised loss and a margin of 0.1 is applied in the ranking loss. The implementation follows the official CutMix and MixUp repositories, using timm’s ViT pretrained on ImageNet-21k.

\subsection{Experiment Results}
\label{ssec:results}

Table 1 summarizes the performance of ViT-Small and ViT-Base under three different sampling strategies for augmentation: Patient+Exam-level, Exam-level, and Image-level. Across both backbones, incorporating Patient+Exam-level sampling consistently yields the highest AUROC, AUPRC, and external C-index. For example, ViT-Small achieves 67.22\% AUROC and 63.79\% AUPRC with the Patient+Exam-level strategy, compared to 63.54\% and 58.25\% under the Image-level sampling. Similar trends are observed with ViT-Base, where the Patient+Exam-level sampling improves the external C-index to 64.98\%, significantly outperforming the Image-level sampling (49.62\%).

\subsection{Ablation Studies}
\label{ssec:ablation}

\begin{table}[t]
    \centering
    \setlength{\tabcolsep}{4pt}
\begin{tabular}{llcccc}
    \toprule
    Metric & Sampling & No Aug & MixUp & CutMix & \makecell{CutMix+\\MixUp} \\
    \midrule
    \multirow{3}{*}{AUROC} 
      & P+E &  - & 65.56 & 66.03 & \textbf{67.22} \\
      & E         &  -  & 64.98 & 66.06 & 65.38 \\
      & I        & 63.93  & 62.44 & 64.59 & 63.54 \\
    \midrule
    \multirow{3}{*}{AUPRC} 
      & P+E &  -   & 60.59 & 61.22 & \textbf{63.79} \\
      & E         & - & 60.92 & 60.98 & 60.46 \\
      & I        &  58.76      & 57.29 & 59.83 & 58.25 \\
    \bottomrule
\end{tabular}
    \caption{Results on the internal test set using ViT-S. Comparison of no mixed-sample augmentation (No Aug), MixUp, CutMix, and CutMix+MixUp in terms of AUROC and AUPRC under different sampling strategies. P+E: Patient+Exam, E: Exam-only, I: Image-level mixing (no patient/exam constraint).}
    \label{tab:results}
\end{table}

\textbf{Comparison of CutMix+MixUp, MixUp, CutMix, and without mixed sample augmentation.} We conduct an ablation study comparing MixUp, CutMix, CutMix+MixUp and without mixed sample augmentation, and results on the Alzeye test set using a ViT-Small model are summarized in Table 2.
Applying either MixUp or CutMix individually at the Patient+Exam-level or Exam-level consistently yields improved performance compared to the conventional Image-level approach, suggesting that preserving patient-specific attributes is important in systemic disease prediction. The performance gap between Patient+Exam and Exam-level sampling is small when either CutMix or MixUp is applied alone, but their gap becomes significantly larger when CutMix and MixUp are jointly applied, indicating Patient+Exam level sampling benefits more from strong augmentation.
Furthermore, Figure 3 shows that the Exam-level sampling strategy converges rapidly, reaches an early plateau, and subsequently exhibits signs of overfitting due to its inherently limited combination space. This suggests that Patient+Exam-level sampling not only enhances predictive performance by better preserving patient-specific attributes but also mitigates overfitting by increasing the diversity of mixing combinations compared to Exam-level sampling.

\textbf{Effects of relative supervision in Patient-level sampling.} In Patient+Exam-level sampling, we compare relative supervision using pairwise ranking loss with direct label supervision using cross-entropy loss. For the latter, the labels are defined as a linear combination of the labels from two different exams, following the standard CutMix and MixUp. The corresponding loss function for this direct supervision is formulated as: \[
\mathcal{L} = \mathcal{L}_{\text{sup}}\!\left(y_{t_i}, \hat{y}_{t_i}\right)
+ \mathcal{L}_{\text{sup}}\!\left(y_{\text{mix}}, \hat{y}_{\text{mix}}\right)
\]

Table 3 summarizes the results, showing that relative supervision based on temporal order outperforms direct supervision using linearly combined labels, highlighting the ambiguity in defining labels when mixing samples from different time points.

\begin{table}[t]
    \centering
    \begin{tabular}{llcc}
\toprule
\multicolumn{1}{c}{} & \multicolumn{1}{c} {Method} & AUROC & AUPRC \\
\midrule
\multirow{2}{*}{ViT small} & Relative Sup & 67.22 & 63.79 \\
                           & Direct label Sup   & 65.31 & 61.01 \\
\midrule
\multirow{2}{*}{ViT base}  & Relative Sup & 67.29 & 67.81 \\
                           & Direct label Sup   & 65.60 & 60.68 \\
\bottomrule
\end{tabular}
    \caption{Comparison between relative supervision and direct label supervision on internal test set.}
    \label{tab:results}
\end{table}

\section{conclusion}
\label{sec:typestyle}

In this work, we revisit mixed sample data augmentation in the context of systemic disease prediction using retinal imaging and propose a hierarchical sampling strategy, \textbf{Oculomix}, that leverages the patient--exam-image hierarchy to preserve clinically relevant attributes.

Experiments on both internal and external datasets show that the proposed sampling strategy consistently improves AUROC, AUPRC, and C-index. Moreover, through comprehensive ablation studies, we demonstrate the inherent ambiguity and difficulty in defining mixed labels across exams and validate the need to preserve patient-specific attributes.

\section{Acknowledgments}
\label{sec:acknowledgments}

The research was supported by the National Institute for Health and Care Research (NIHR) Biomedical Research Centre based at Moorfields Eye Hospital NHS Foundation Trust and UCL Institute of Ophthalmology. The views expressed are those of the author(s) and not necessarily those of the NHS, the NIHR or the Department of Health and Social Care.

\bibliographystyle{IEEEbib}
\bibliography{refs}

\end{document}